\documentclass[letterpaper, 10 pt, conference]{ieeeconf}  

\IEEEoverridecommandlockouts                              

\usepackage{bbding}

\usepackage{graphics} 
\usepackage{epsfig} 
\usepackage{mathptmx} 
\usepackage{times} 
\usepackage{amsmath} 
\usepackage{amssymb}  
\usepackage{xcolor} 
\usepackage{booktabs}
\usepackage{tikz}
\usetikzlibrary{arrows.meta,positioning,fit,calc}
\title{\LARGE \bf Co-Speech with You: Training-Free Personalization of Robot Co-Speech Gestures
}

\author{
Bosong Ding$^{1}$, Selma Ancel$^{1}$, Giacomo Spigler$^{1}$, Murat Kirtay$^{1}$
\thanks{$^{1}$The authors are with the Department of Cognitive Science and Artificial Intelligence, Tilburg University, Tilburg, Netherlands.
{\tt\small {b.ding\_3, S.J.Ancel, g.spigler, m.kirtay}@tilburguniversity.edu}}%
}

\begin{document}

\maketitle
\thispagestyle{empty}
\pagestyle{empty}

\begin{abstract}
Personal robots should adapt their co-speech gesture style to a new user without requiring model retraining. We present a training-free personalization pipeline that combines a frozen audio-conditioned diffusion prior with a gesture style encoder and lightweight conditioning adapters. The encoder is first trained to discriminate speaker identities and then jointly refined with the adapters using the diffusion objective, enabling a reusable style embedding to be extracted from approximately 10 seconds of enrollment motion through a single forward pass. To support this setting, we also release a Quest~3 capture application and a dataset of spontaneous co-speech motion from ten participants. We evaluate the system on held-out speakers using Style Recognition Accuracy (SRA) and Fr'echet Gesture Distance (FGD) to measure personalization and motion quality. Our approach improves SRA from 27.6\% for the frozen prior to 69.5\% while preserving motion quality (FGD 34.2 versus 34.8), and replacing the enrollment embedding with another person's reduces SRA to 11.4\%. The generated gestures are retargeted to a physical NAO robot, and this improvement also transfers to the robot deployment setting, where speech is synthesized using five TTS voices, retaining 67.3\% SRA.
\end{abstract}

\section{INTRODUCTION}

People do not gesture in a generic way during a conversation. Given the same speech script, different speakers vary in how often they gesture, how broadly they use space, how quickly they move, and which hand they favor \cite{neff2008gesture, ginosar2019learning}. These features are part of a person's nonverbal style that plays a potent role during interaction. Co-speech gestures are communicative body movements, especially of the hands and arms, that accompany speech \cite{nyatsanga2023comprehensive}. A co-speech gesture generator that ignores nonverbal style can produce motion that is synchronized with speech, yet still feels unlike the person it is meant to represent. This is a limiting factor for personal robots, where gesture shapes how the robot is perceived: a robot that accompanies, assists, or represents a particular user should be able to reflect that user's gestural style rather than reproduce the average behavior of a training dataset \cite{salem2012generation}. This paper frames the following research question: can a robot gesture like a new user from observing their natural co-speech motion?

\begin{figure}[h]
    \centering
    \begin{minipage}{\linewidth}
        \centering
        \includegraphics[width=\linewidth]{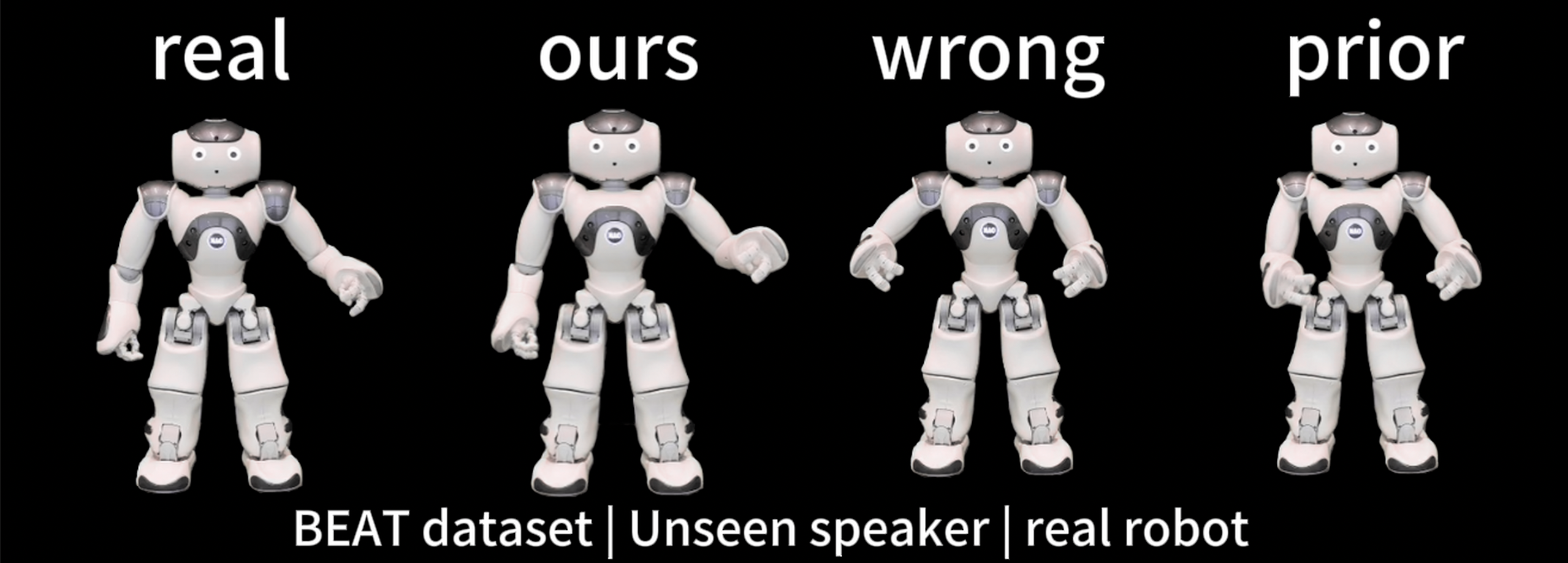}
    \end{minipage}%
    
    \begin{minipage}{\linewidth}
        \centering
        \includegraphics[width=\linewidth]{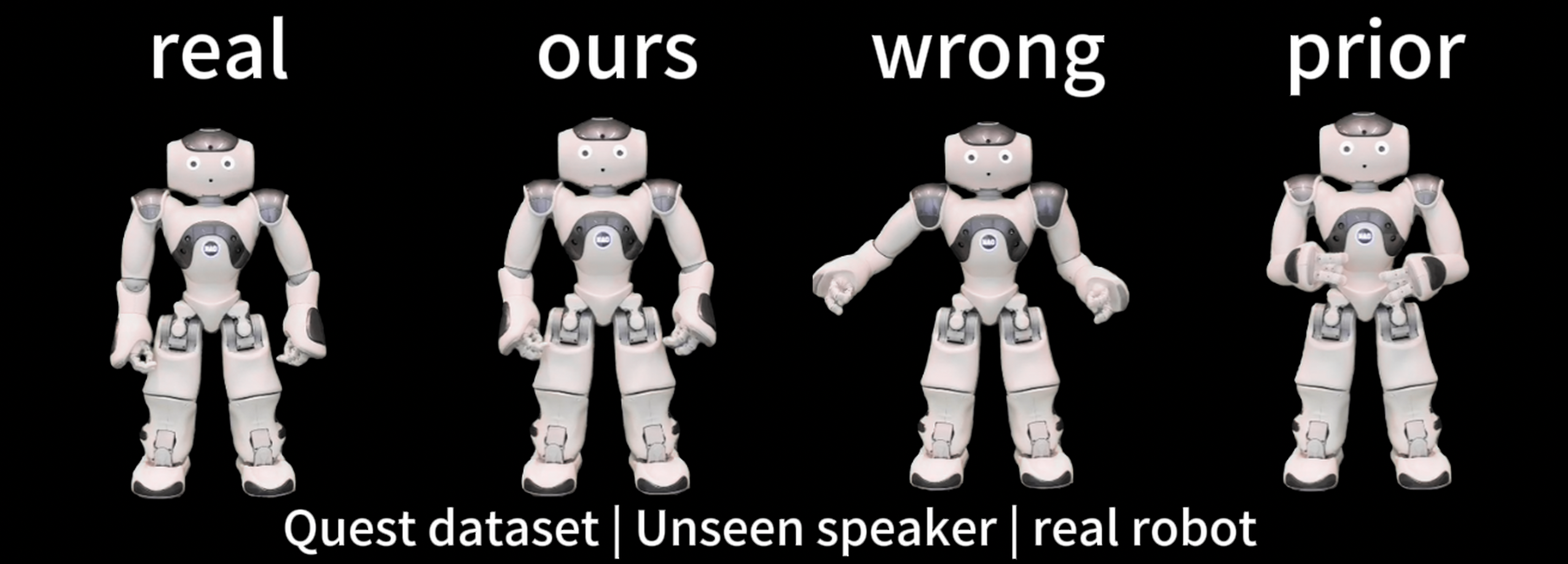}
    \end{minipage}
    
    \caption{Personalized co-speech gestures executed on a real NAO robot for
    speakers never seen in training. For the same speech, each row compares the
    speaker's own motion (real), our personalized output (ours), output
    conditioned on another person's style embedding (wrong), and the frozen
    diffusion prior without personalization (prior). Top: an unseen BEAT
    speaker. Bottom: an unseen participant from our Quest~3 dataset.}
    \label{fig:naodemo}
\end{figure}
To address this question, we propose a personalized co-speech generation system using Quest~3 virtual reality headset. We then employed the system on a humanoid robot that pairs an audio-conditioned diffusion model with a gesture style encoder.
To do this, we first initialize the encoder by discriminating among 1,214 training identities, then jointly
refine it with a small style conditioning while keeping the diffusion prior frozen.

At deployment, the encoder maps twelve 2.27-s windows (9.6\,s total)
of a new person's tracked motion to a style embedding; enrollment requires only a forward pass and no
person-specific optimization. In our setting, the person is enrolled through a Quest~3
headset rather than the robot's own camera. We used this approach because monocular
pose estimates can corrupt the fine temporal variation used to identify gesture
style. Finally, the generated motion is retargeted to a NAO robot.


We evaluate speakers using Style Recognition
Accuracy (SRA) for identity and Fr\'echet Gesture Distance (FGD) for realism~\cite{yoon2020speech,kucherenko2024evaluating}.
On the seven-person BEAT/Trinity test set, a short enrollment recording raises SRA
from $27.6\%$ for the frozen prior to $69.5\%$, against $14.3\%$ chance level,
while retaining comparable FGD (34.2 versus 34.8).
Substituting another
person's style embedding reduces SRA to $11.4\%$, showing that the gain follows the
enrollment identity rather than style guidance alone. On ten unseen people captured with a Quest~3
headset, SRA increases from $10.0\%$ to $32.3\%$ and FGD decreases from $39.4$
to $30.5$. With five synthetic TTS voices, target SRA remains $67.3\%$ and
falls to $1.6\%$ under a wrong-person style embedding. 
Overall, we show that our co-speech system adds target-specific gestural identity without user-specific optimization
and maps the resulting motion to a NAO robot. To support shared evaluation, we report these metrics across encoders, baselines, judges, and seeds \cite{wolfert2022review, kucherenko2024evaluating}.
  
  
This paper presents the following contributions: 
  \begin{enumerate}
  \item An optimization-free enrollment mechanism that controls an
  unseen person's gesture identity while preserving the frozen diffusion
  prior's motion quality.
  \item A publicly available co-speech recording software and a dataset of unscripted, spontaneous everyday co-speech behavior.
  \item An end-to-end real-robot implementation to show
  person-specific motion remains recognizable, visible, and executable after
  retargeting to the robot's reduced kinematic chain.
  \end{enumerate}

\section{RELATED WORK}
  \label{sec:related}

  \subsection{Speech-driven gesture generation}
   Speech provides semantic and rhythmic cues for gesture, but it does not specify a single correct non-verbal movement, as the same speech can be accompanied by many plausible gestures. To learn the range of motions that may accompany speech, data-driven methods use recurrent, adversarial, and variational architectures, conditioning gesture generation on information such as audio, text, an initial pose, or speaker identity \cite{yoon2020speech,liu2022learning,nyatsanga2023comprehensive}. More recent approaches use normalizing flows and diffusion models to better represent the one-to-many relationship between speech and gesture \cite{henter2020moglow,zhu2023taming}. However, methods that represent identity using a categorical speaker label are limited to speakers observed during training and cannot directly personalize gestures for a previously unseen user.

  \subsection{Personalization and example-based conditioning}
  Early approaches to personalized gesture generation trained a separate model for each speaker
  \cite{neff2008gesture,ginosar2019learning,habibie2021learning,yi2023generating}. Although this strategy can capture speaker-specific motion patterns, it
  requires substantial training data and optimization for every new person. Later approaches reduced this requirement. For example, DiffGAN adapts to a new speaker using
  approximately two minutes of motion data \cite{ahuja2022low}, while C-DiffGAN supports continual adaptation as new speakers are introduced
  \cite{ahuja2023continual}. Nevertheless, these methods still optimize model parameters for each user. Other approaches encode identity through a learned
  categorical speaker representation \cite{ahuja2020style}. Yet, in these studies, representations are limited to speakers included during training and cannot directly describe a previously unseen person.

  To overcome this limitation, example-conditioned methods were proposed to infer style from a reference motion clip, avoiding user-specific optimization at test time. ZeroEGGS demonstrates this approach by encoding a motion example into a style representation that controls gesture generation \cite{ghorbani2023zeroeggs}. However, it is
  trained on one actor performing 19 instructed styles, which differs from learning the gestural habits that distinguish individuals. Other systems control gesture style through predefined labels or textual prompts \cite{yang2023diffusestylegesture,ao2023gesturediffuclip,chen2025motion}.
  These approaches provide control over acted or verbally described styles, e.g., angry, old, energetic, but they do not establish whether a model can capture the everyday co-speech behavior of a previously unseen person from a brief, natural motion sample.

  \subsection{Co-speech gesture on physical robots}
  Human motion must be retargeted to a robot with fewer degrees of freedom while
  respecting its joint limits and executable motion ranges
  \cite{ding2024imitation}.Deploying co-speech gestures on a physical robot introduces constraints that are not present in virtual settings, e.g., character animation. Human motion must be retargeted to
  a robot with fewer degrees of freedom while respecting its joint limits and executable motion ranges. 
  
  Yoon et al. generate co-speech motion and analytically retarget it to the NAO robot \cite{yoon2019robots}. TED-Culture extends this direction by mapping
  culturally conditioned gestures from a DiffGesture-based system to the NAO robot, using separately trained language models to provide linguistic context
  \cite{shen2025ted}. Other work has generated humanoid robot gestures conditioned on emotional states \cite{huang2025emotion}. These systems control
  general, cultural, or emotional properties of robot motion, but they do not personalize gestures from a previously unseen
  individual. They also do not measure how much of that individual’s gestural identity remains recognizable after human motion is transferred to the robot’s reduced kinematic structure.

Taken together, these studies do not resolve the main challenge: a physical robot that can enroll a previously unseen person from a brief co-speech motion sample, personalize without per-user optimization, and retain that person’s recognizable gestural identity after retargeting.

  \section{Motion Capture with Quest~3}
  \label{sec:capture}
Personalizing gestures for a previously unseen user requires a practical way to observe that user’s co-speech motion. Co-speech datasets commonly obtain motion either from marker-based motion capture or from monocular video, as summarized in Table~\ref{tab:dataset_comparison_simplified}. Motion-capture (mocap) systems provide
  accurate trajectories but require specialized hardware and controlled data
  collection. Consequently, even BEAT dataset, one of the largest mocap corpora,
  contains only 30 speakers. Video-based reconstruction scales to more
  speakers, but frame-wise pose errors can introduce implausible limb extensions
  and high-frequency jitter. This trade-off is
  especially problematic for personalization: broad identity coverage is needed
  to learn inter-person variation, while clean trajectories are needed to
  generate executable robot motion.
  
  We quantify capture smoothness using mean jerk. For a direction-vector sequence
  $\mathbf{X}=\{\mathbf{x}_1,\ldots,\mathbf{x}_T\}$, we define
  \begin{equation}
    j(\mathbf{X})=\operatorname*{mean}_{t}\left\|\Delta^3\mathbf{x}_t\right\|_2 ,
  \end{equation}
  where $\mathbf{x}_t$ is the skeleton direction vector at frame $t$, $\Delta^3$ is the third temporal difference, and lower values indicate
  smoother motion. TED-Ex motion reconstructed from video has a jerk of \textbf{$5.13$},
  compared with \textbf{$0.67$} for suit-based BEAT and \textbf{$0.84$} for our Quest~3 capture.
  Thus, Quest~3 is much closer to wearable suit-based motion capture than to
  video-reconstructed one under this smoothness measure. Complementary simultaneous validation against laboratory motion capture found high concordance and consistency for most Quest~3 upper-limb kinematic variables
  \cite{campos2026quest}

Together, these results support Quest~3 as a high-quality, portable source of upper-limb motion. To our knowledge, this is the first reported use of Quest~3 for co-speech gesture capture. The headset directly tracks the upper body, hands, and fingers, reducing the occlusion and camera-motion errors common in monocular robot camera capture. For privacy concerns, the application stores only motion coordinates and synchronized audio, without recording any RGB video. Its recordings are resampled to 15Hz and converted to the same mean-subtracted 126-D direction-vector representation used by our model.
\begin{table}[h]
        \centering
        \caption{Open-source co-speech gesture datasets.}
        \begin{tabular}{lccccc}
        \hline
        Dataset & Duration (hours) & Speakers & Gesture\\
        \hline
        Trinity \cite{ferstl2018investigating} & 4 & 1  & \textbf{mo-cap} \\
        SCG \cite{habibie2021learning} & 33 & 6  & video \\
        TED-Ex \cite{liu2022learning} & 100.8 & 1,764  & video \\
        ZeroEGGS \cite{ghorbani2023zeroeggs} & 2 & 1 & \textbf{mo-cap} \\
        BEAT \cite{liu2022beat} & 35 & 30 & \textbf{mo-cap} \\
        TalkSHOW \cite{yi2023generating}& 26.9 & 4 & video \\
        \hline
        \end{tabular}
        
        \label{tab:dataset_comparison_simplified}
        \end{table}

  Using this application, we acquired a dataset designed to evaluate
  personalization during natural, everyday speech. This setting is important
  because personal gesture style appears in subtle habits---such as gesture
  frequency, scale, speed, and handedness---that may be suppressed or replaced
  by exaggerated, task-driven behavior in a motion capture studio. Quest~3 passthrough allows
  participants to remain aware of their surroundings and record in familiar,
  comfortable environments, in the office and on streets. In our setting, each participant spoke in English for approximately five minutes about a topic
  or story that interested them. The data collection used no script and prohibited
  instructed gestures or acted emotions, so the recordings reflect habitual
  behavior in a spontaneous monologue to a visible listener. Each participant was recorded in a single session on a single day, so gestural style does not vary across recording days within a participant. Participants are of mixed ethnic backgrounds and
  consented to collection and controlled sharing after anonymization and voice
  transformation. The resulting dataset contains 51.2 minutes of co-speech motion from 10 individuals. Manual review excludes headset wearing and removal, non-conversational intervals, and segments affected by technical issues. Because the dataset contains
  identifiable voice recordings, it is distributed under controlled access and is available to researchers upon approval of a data-access request.

 \begin{figure}[h]
    \centering
    \includegraphics[width=\linewidth]{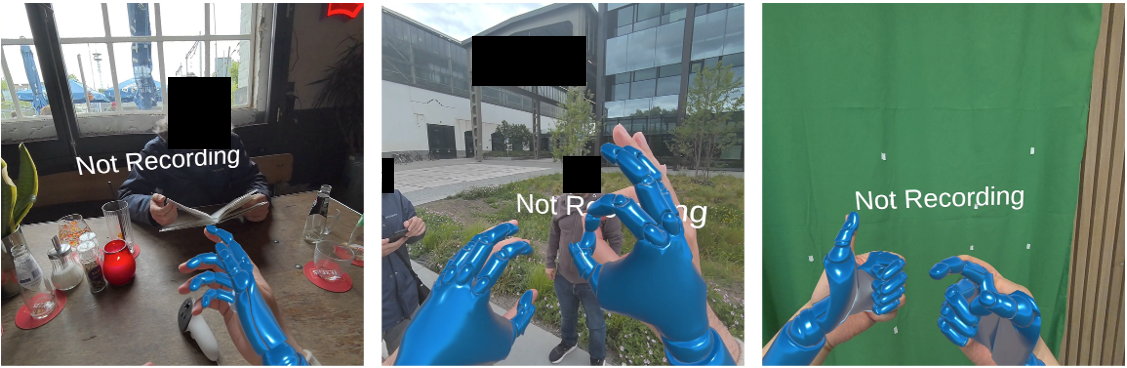}
    \caption{Quest~3 upper-body and hand capture interface.}
    \label{fig:quest_demo}
\end{figure}
\begin{figure*}[t]
  \centering
  \includegraphics[width=\linewidth]{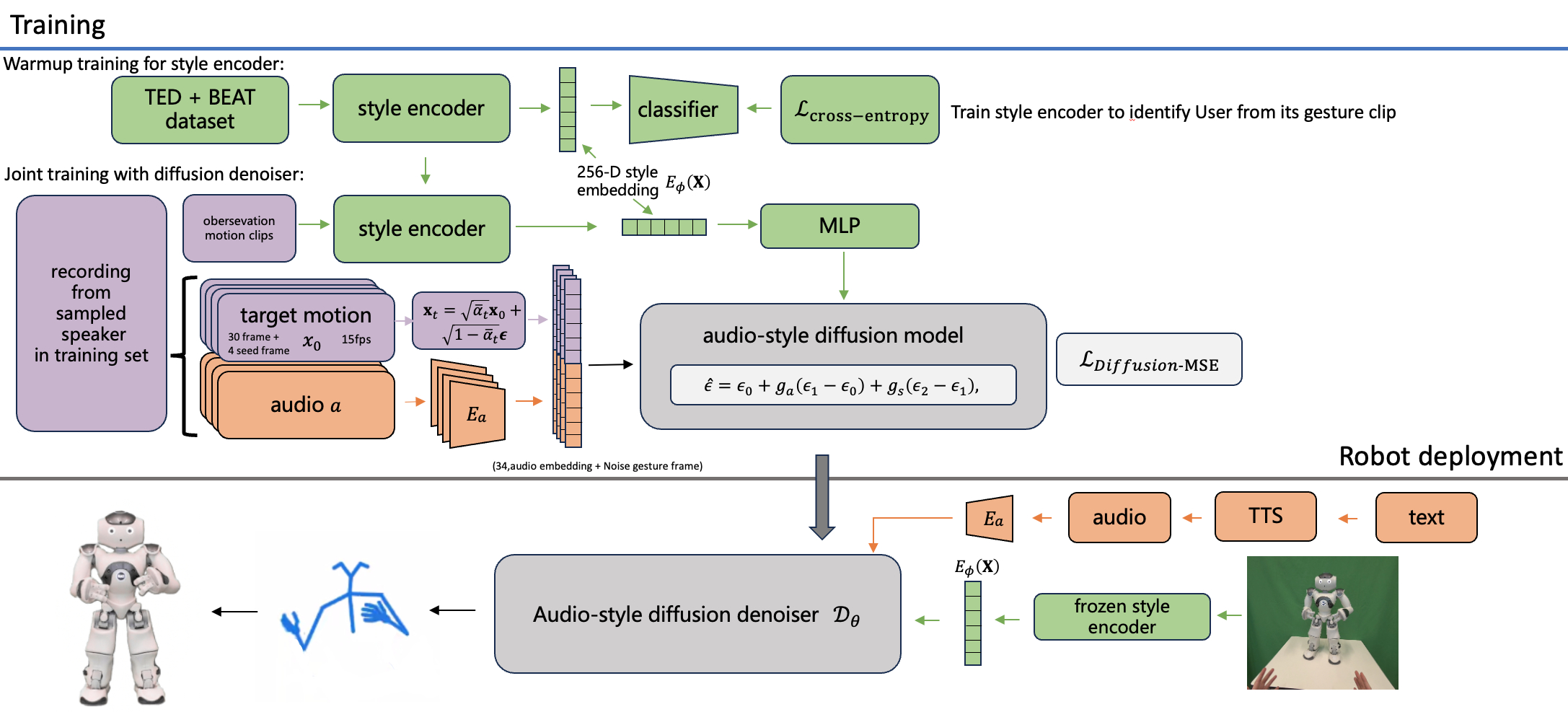}
  \caption{Overview of the proposed pipeline. Top: the style encoder is
  initialized by separating TED-Ex and BEAT identities, then jointly refined with
  zero-initialized style adapters while the diffusion prior remains frozen.
  Bottom: at deployment,
  a large language model (LLM) and TTS provide speech, enrollment motion provides a target style
  embedding, and the shared denoiser combines audio and style through guidance
  weights $g_a$ and $g_s$. The generated direction vectors are retargeted
  analytically to the NAO.}
  \label{fig:main_pipeline}
\end{figure*}

\section{Diffusion model for Personalization}
\label{sec:method}
This section introduces our diffusion model for personalized co-speech motion. We first train two components separately: a style encoder to discriminate
  speaker identities and an audio-conditioned diffusion model on BEAT to learn a general motion prior. We then freeze the diffusion model and jointly
  refine the style encoder with a lightweight style adapter that modulates the denoiser. At deployment, the encoder extracts a reusable style embedding from a
  short enrollment recording, enabling a new user to be enrolled without updating the model. Figure~\ref{fig:main_pipeline} summarizes training and
  deployment.

\subsection{Input Data}
We select TED-Ex\cite{liu2022learning} and BEAT\cite{liu2022beat} datasets because they provide complementary properties. Among the datasets summarized in Table~\ref{tab:dataset_comparison_simplified}, TED-Ex provides the broadest speaker coverage, whereas BEAT is the largest mo-cap dataset. TED-Ex provides identity diversity for representation learning, whereas BEAT provides clean motion-capture trajectories for gesture generation.

All motion is represented using mean-subtracted 126-D joint direction vectors. We resample the trajectories to 15Hz and divide them into 34-frame (2.27\,s) windows with a stride of 10 frames. At enrollment, the style embedding is estimated from 12 consecutive windows, which together span approximately 9.6\,s of source motion. We select this approximately 10-s observation duration as a practical balance between capturing gestural variation and limiting the amount of required enrollment data. A 10-s observation has been used to evaluate impressions of speaking robots and affective robot gestures \cite{gurung2024uncanny}, while conversational gesture sequences of similar duration produce measurable differences in perceived personality \cite{smith2017understanding}. Comparable durations also appear in zero-shot gesture generation studies \cite{ghorbani2023zeroeggs,fares2023zero}.

After preprocessing and minimum-duration filtering, 1,290 of TED-Ex's 1,764 speakers provide sufficient motion for this protocol. We reserve 100 as unseen identities for encoder evaluation and use the remaining 1,190 for training. Together with 24 of BEAT's 30 speakers, they form the 1,214 identities used for discriminative style-encoder initialization. The diffusion prior and subsequent joint refinement are trained exclusively on BEAT's motion-capture trajectories, preventing video-reconstruction noise from entering the generation targets. For unseen-person evaluation, we reserve six BEAT speakers and one Trinity speaker~\cite{ferstl2018investigating} from both encoder and generator training; like BEAT, Trinity is motion-captured and non-acted, yielding a seven-speaker BEAT/Trinity test set. We select Trinity as the only other mocap, non-acted source in Table~\ref{tab:dataset_comparison_simplified}; ZeroEGGS is acted~\cite{ghorbani2023zeroeggs} and the remaining datasets are video-reconstructed (Section~\ref{sec:capture}). Our ten-person Quest~3 dataset serves as a second test set whose participants and styles are both unseen during training.

\subsection{Style Encoder}
Personalization depends on encoding gestural properties that remain characteristic
of a person across speech, such as movement amplitude, speed, and handedness. Both the training objective and temporal architecture determine which of these properties are retained. We compare reconstruction, variational, recurrent, and discriminative alternatives and select the
ZeroEGGS attention architecture \cite{ghorbani2023zeroeggs} with a speaker
classification objective, which gives the strongest separation of unseen
identities in our encoder ablation in section \ref{abla-encoder}.

The encoder applies two temporal convolutions followed by one feed-forward
  Transformer (FFT) attention block and temporal mean pooling. It maps each
  \(34\times126\) motion window \(\mathbf{X}\) to a 256-D style embedding
  \(\mathbf{s}=E_\phi(\mathbf{X})\in\mathbb{R}^{256}\), where \(\phi\) are the encoder parameters. We first optimize \(E_\phi\) with cross-entropy (CE) over the
  1,214 training identities. The resulting checkpoint initializes the
  deterministic encoder used in the second training stage, where it is jointly
  refined with the style adapter through the diffusion objective. At deployment, the refined encoder remains fixed and produces a style embedding without requiring optimization.

\subsection{Diffusion Model}
Unlike personalizing an animated character\cite{fares2023zero,ghorbani2023zeroeggs}, a physical robot directly executes the generated motion. \textbf{Personalization therefore cannot come at the expense of plausible and stable behavior}. A style embedding inferred from a short enrollment recording may be imperfect or weakly expressed; allowing it to dominate generation could result in motion that could damage the hardware. We therefore use the style embedding as a controlled modulation of a frozen audio-conditioned diffusion prior. The prior anchors generation in general co-speech motion, while the style adapter introduces person-specific variation without modifying the denoiser weights. The diffusion model perfectly supports this separation by modeling the one-to-many mapping from speech to motion and allowing the influence of each condition to be controlled during iterative denoising \cite{zhu2023taming}.

The prior operates on 34-frame direction-vector windows and uses a 500-step denoising process. A convolutional audio encoder supplies speech features aligned with frames to a Transformer denoiser with a 512-D hidden representation and eight blocks. Let $\mathbf{x}_0$ be a clean motion window, $\epsilon\sim\mathcal{N}(0,I)$, and $\mathbf{x}_t=\sqrt{\bar{\alpha}_t}\mathbf{x}_0+\sqrt{1-\bar{\alpha}_t}\epsilon$, where $t$ is the diffusion timestep and $\bar{\alpha}_t$ is the cumulative noise schedule. The model is trained with the standard
noise-prediction objective
\begin{equation}
  \mathcal{L}_{\mathrm{diff}} =
  \mathbb{E}_{t,\mathbf{x}_0,\epsilon}
  \left[\left\|\epsilon-
  \epsilon_\theta(\mathbf{x}_t,t,\mathbf{a})\right\|_2^2\right],
\end{equation}
where $\theta$ are the denoiser parameters and $\mathbf{a}$ denotes the audio and prefix context.

We freeze this prior and introduce a style-conditioning path. A two-layer multilayer perceptron (MLP)
projects the 256-D style embedding $\mathbf{s}$ to 128 dimensions. The projected embedding supplies
an additive bias after the denoiser input projection and adaptive normalization
parameters at every Transformer block:
\begin{equation}
  \widetilde{\mathbf{h}}_\ell =
  \mathbf{h}_\ell\odot(1+\boldsymbol{\gamma}_\ell(\mathbf{s}))
  +\boldsymbol{\beta}_\ell(\mathbf{s}),
\end{equation}
where $\mathbf{h}_\ell$ is the hidden state at Transformer block $\ell$ and $\boldsymbol{\gamma}_\ell$, $\boldsymbol{\beta}_\ell$ are adaptive scale and shift heads conditioned on the style embedding $\mathbf{s}$.
All additive, scale, and shift heads are initialized from scratch, so the
personalized network is initially identical to the frozen prior. We then
optimize the adapter and the deterministically initialized style encoder jointly
with $\mathcal{L}_{\mathrm{diff}}$ for 20 epochs; all diffusion-prior parameters
remain fixed. This one-time joint refinement aligns the style embedding with the generation objective without updating the prior or requiring any optimization for an enrolled test user. On 10\% of training examples, the style embedding is replaced by a learned null style embedding to enable classifier-free style guidance to preserve the general good quality motion prior.

At sampling time, we compose the prior's audio guidance with the new style guidance. For one denoising step, let $\epsilon_0$, $\epsilon_1$, and $\epsilon_2$ denote the denoiser's noise predictions conditioned on (null audio, null style embedding), (audio, null style embedding), and (audio, target style embedding), respectively. We use
\begin{equation}
  \widehat{\epsilon} =
  \epsilon_0 + g_a(\epsilon_1-\epsilon_0)
  + g_s(\epsilon_2-\epsilon_1),
\end{equation}
with audio guidance $g_a=1.15$. The null-style branch preserves the
speaker-agnostic prior, while $g_s$ controls how strongly the generated motion
follows the style embedding. We use $g_s=6$ for the headline comparisons. All
evaluation conditions use a zero-motion prefix, so no target motion is
copied into generation.

\subsection{Robot Embodiment}
Following prior NAO robot co-speech systems \cite{yoon2019robots,shen2025ted}, we map the generated human motion to the robot by computing joint angles directly from the limb directions. Each upper-arm direction determines ShoulderPitch and ShoulderRoll, while the corresponding forearm direction determines ElbowYaw and ElbowRoll. We smooth each joint trajectory using a second-order Savitzky--Golay filter with a five-frame window. Five frames is the shortest symmetric window that provides non-trivial smoothing for a quadratic fit and corresponds to only \(0.33\,\mathrm{s}\) at 15\,Hz, thereby limiting temporal averaging of the generated motion. The resulting angles are then clamped to the documented NAO robot joint limits.

Because the motion prior is trained on motion-capture trajectories rather than video-reconstructed poses, we keep robot-side post-processing minimal: the same minimum smoothing filter and joint-limit clamp are \textbf{applied to all generated motion}, without the joint-specific jitter corrections used in a video-trained NAO robot pipeline \cite{shen2025ted}. This fixed mapping allows the application study to evaluate personalization produced by the generator: Figure~\ref{fig:naodemo} shows the resulting motions on the physical robot, while Table~\ref{tab:robot_sra} quantifies how much person-specific information remains after conversion to the NAO's eight-joint representation.

\section{Results and Ablation Study}
\begin{table*}[t]
  \centering
  \caption{Five-seed personalization results.
  The TTS column reports seven-way target SRA, macro-averaged across five
  voices. All personalized conditions use $g_s{=}6$. Values are mean $\pm$
  sample standard deviation.}
  \label{tab:main_personalization}
  \small
  \begin{tabular}{l cc cc c}
    \toprule
    & \multicolumn{2}{c}{Unseen BEAT/Trinity speakers}
    & \multicolumn{2}{c}{Quest~3}
    & \multicolumn{1}{c}{TTS audio} \\
    \cmidrule(lr){2-3}\cmidrule(lr){4-5}\cmidrule(lr){6-6}
    Condition
    & SRA (\%) $\uparrow$ & FGD $\downarrow$
    & SRA (\%) $\uparrow$ & FGD $\downarrow$
    & Target SRA (\%) $\uparrow$ \\
    \midrule
    Chance
    & $14.3$ & {--}
    & $10.0$ & {--}
    & $14.3$ \\
    Real-motion ceiling
    & $86.7 \pm 10.1$ & {--}
    & $68.7 \pm 2.7$ & {--}
    & {--} \\
    Frozen prior
    & $27.6 \pm 2.5$ & $34.8 \pm 6.6$
    & $10.0 \pm 0.4$ & $39.4 \pm 4.1$
    & {--} \\
    Wrong style embedding ($g{=}6$)
    & $11.4 \pm 4.4$ & $43.8 \pm 8.4$
    & $4.5 \pm 2.2$ & $49.1 \pm 14.1$
    & $1.6 \pm 1.7$ \\
    Ours ($g{=}6$)
    & $\mathbf{69.5 \pm 11.1}$ & $\mathbf{34.2 \pm 5.0}$
    & $\mathbf{32.3 \pm 8.5}$ & $\mathbf{30.5 \pm 4.5}$
    & $\mathbf{67.3 \pm 10.8}$ \\
    \bottomrule
  \end{tabular}
\end{table*}
\label{sec:results}
This section evaluates whether our proposed model gives the robot a recognizable person-specific gestural identity\textbf{ without sacrificing the motion quality} of the frozen diffusion prior. Our objective is therefore not to maximize identity recognition alone, but to obtain high SRA while maintaining FGD at or below the unpersonalized prior. We first define these complementary metrics, then test the resulting operating point across unseen speakers from BEAT and our Quest~3 dataset, and synthetic voices. We then quantitatively measure the robot retargeting loss on the personalization, and finally model ablations.
\subsection{Metrics}
  \label{sec:metrics}

  Co-speech gesture is inherently one-to-many, so frame-wise error against a
  single recording may penalize plausible alternative gestures. We therefore use
  \textbf{Fr\'echet Gesture Distance (FGD)}, a widely used measure of the
  distributional distance between generated and real motion in the latent space
  of a fixed motion autoencoder~\cite{yoon2020speech,zhu2023taming}. Lower FGD
  indicates a closer distributional match. We interpret FGD as a
  proxy for motion quality, rather than an absolute naturalness score or a value
  that can be compared across different skeletons and evaluators.

  FGD measures motion realism, not whether the generated motion resembles the intended person. Classifier-based
  \textbf{Style Recognition Accuracy (SRA)} has therefore been used to evaluate whether
  generated gestures express the intended style~\cite{ao2023gesturediffuclip}.
  We adapt this principle to enrolled-person identity: SRA is the fraction of
  generated clips assigned to the correct enrolled person by an independent
  judge trained only on held-in identities. Each candidate is represented by the
  mean judge embedding of their real clips, and generated clips are assigned by
  nearest cosine similarity. Chance accuracy is $1/|\mathcal{C}|$ for candidate
  set $\mathcal{C}$.

  Our objective is consequently to maximize SRA while maintaining FGD at or below
  the frozen prior: the robot should acquire a recognizable identity and
  personality style without sacrificing motion quality even under poor or wrong style guidance.

\subsection{Test-set Results}
Table~\ref{tab:main_personalization} reports our central result: whether the
  final model improves identity recognition for new users without degrading
  motion quality. All evaluation participants are excluded from style-encoder,
  diffusion-prior, and embedding-adapter training. We test seven held-out
  BEAT/Trinity speakers and all ten Quest~3 participants. Person-specific style
  is not perfectly separable even in real motion, because different people may
  gesture similarly. Accordingly, real motion reaches empirical SRA ceilings of
  $86.7\%$ on BEAT/Trinity and $68.7\%$ on Quest~3 rather than $100\%$.

  With style guidance $g_s=6$, our model raises BEAT/Trinity SRA from $27.6\%$
  to $69.5\%$ ($+41.9$ points), while mean FGD remains effectively unchanged
  ($34.8$ versus $34.2$). On Quest~3, SRA rises from $10.0\%$ to $32.3\%$
  ($+22.3$ points), while FGD decreases from $39.4$ to $30.5$. The model
  therefore improves personalization without a measurable loss of motion quality
  for unseen motion-capture speakers, and preserves this trade-off under unseen
  consumer capture.

  The wrong-embedding control verifies that these gains are driven by the
  enrolled style embedding. Replacing it with a deliberately mismatched
  other-person embedding reduces target SRA to $11.4\%$ on BEAT/Trinity and
  $4.5\%$ on Quest~3, near or below the corresponding chance levels of $14.3\%$
  and $10.0\%$. The resulting motions are instead assigned to the embedding
  owner in $74.9\pm7.8\%$ and $44.1\pm11.5\%$ of cases. Thus, the embedding
  controls whose gesture style is generated rather than merely strengthening
  identity cues already present in the audio.

  Because the intended application is a speaking robot, performance with
  target-person speech alone is insufficient: deployment audio will typically
  use a synthetic voice that contains no acoustic cues from the
  enrolled person. Across five synthetic neutral voices, the enrolled embedding retains
  $67.3\pm10.8\%$ target SRA. Substituting a wrong-person embedding reduces
  target SRA to $1.6\pm1.7\%$ and redirects $79.7\pm8.3\%$ of the motions to
  the embedding owner. We do not report FGD for this condition because the
  synthetic utterances have no corresponding target-person real-motion
  distribution.

\begin{figure}[t]
  \centering
  \includegraphics[width=\columnwidth]{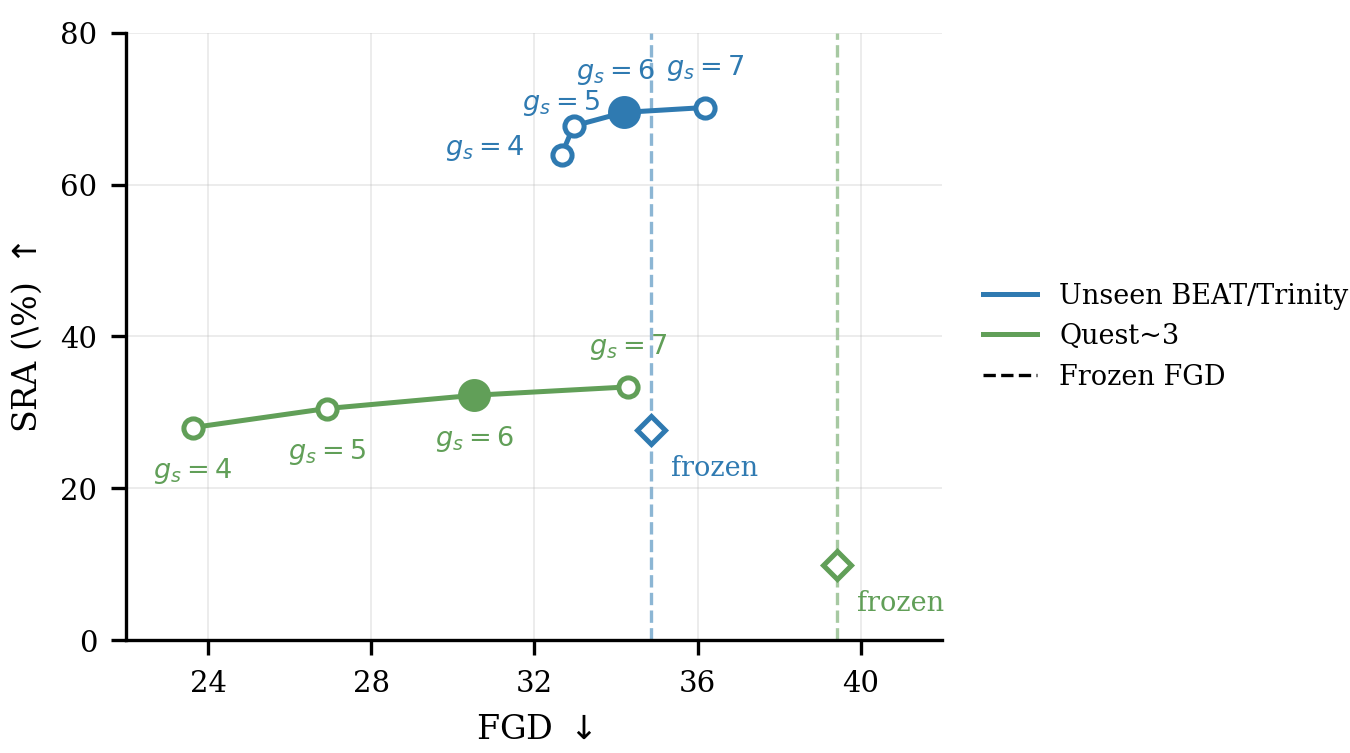}
  \caption{Personalization--motion-quality trade-off across style-guidance
  strengths $g_s\in\{4,5,6,7\}$ on the BEAT/Trinity and Quest~3 test sets
  (five-seed mean). Each curve connects the four operating points; the dashed
  vertical line marks the frozen prior's FGD for that test set. The selected
  $g_s=6$ point (filled) stays at or below the frozen FGD while maximizing
  SRA. Higher SRA and lower FGD are better.}
  \label{fig:guidance_tradeoff}
\end{figure}

  Figure~\ref{fig:guidance_tradeoff} provides a supporting analysis of the
  operating point. Among $g_s\in\{4,5,6,7\}$, $g_s=6$ achieves the highest SRA
  while keeping mean FGD at or below the frozen prior on both test sets.
  Quest~3 can tolerate stronger guidance, but increasing to $g_s=7$ raises
  BEAT/Trinity FGD from the prior's $34.8$ to $36.2$. We therefore use $g_s=6$
  as a single operating point that preserves motion quality across both domains.

  Together, these results show that the learned style embedding captures the
  person-specific information needed to control gestural identity without
  sacrificing the motion quality of the frozen prior. The lower real-motion
  ceiling and smaller personalization gain on Quest~3 indicate that
  person-specific differences are harder to distinguish in a casual environment. This may reflect subtler everyday gestures than those produced in a
  studio, although capture hardware and population differences are also
  confounded. Quest~3 therefore provides a complementary and more challenging
  test of personalization under everyday conditions.

 \subsection{Robot-Space Personalization and Deployment}

  Human-space results do not by themselves establish robot-level
  personalization, because retargeting compresses each 126-D motion into the
  NAO's eight actuated shoulder/elbow joints and further modifies it through
  smoothing and joint-limit clamping. We therefore test whether person-specific
  information remains in the joint trajectories that are delivered to the
  robot. Real motion, the frozen prior, and the correct- and wrong-embedding
  conditions are all processed using the same analytic mapping, five-frame
  filter, and joint-range clamp.

  To quantify identity in this robot representation, we train a separate
  robot-space SRA judge on eight-joint trajectories retargeted from 1,214 held-in
  TED-Ex and BEAT identities. The judge is frozen before evaluation on the seven
  unseen BEAT/Trinity speakers, with prototypes and evaluation clips obtained
  from disjoint real-motion windows.

  \begin{table}[t]
    \centering
    \caption{Identity recognition after eight-joint NAO robot retargeting.}
    \label{tab:robot_sra}
    \small
    \begin{tabular}{l c}
      \toprule
      Motion supplied to retargeting & Robot SRA (\%) $\uparrow$ \\
      \midrule
      Chance & $14.3$ \\
      Retargeted real motion & $53.8 \pm 14.0$ \\
      Frozen prior & $15.1 \pm 2.6$ \\
      Wrong style embedding ($g_s{=}6$) & $8.5 \pm 4.9$ \\
      Ours ($g_s{=}6$) & $\mathbf{33.5 \pm 5.2}$ \\
      \bottomrule
    \end{tabular}
  \end{table}

  As shown in Table~\ref{tab:robot_sra}, the empirical real-motion ceiling falls
  from $86.7\pm10.1\%$ in the original human-motion representation to
  $53.8\pm14.0\%$ after NAO retargeting. This drop quantifies the identity
  information lost through the robot's limited degrees of freedom, smoothing,
  and joint-range constraints. Within this reduced ceiling, our generated motion
  retains $33.5\pm5.2\%$ SRA, compared with $15.1\pm2.6\%$ for the frozen prior
  and $8.5\pm4.9\%$ for the wrong-embedding condition. Moreover,
  wrong-embedding trajectories are assigned to the embedding owner in
  $40.8\pm11.9\%$ of cases. Person-specific control therefore remains
  quantitatively detectable in the executable robot representation, although
  part of the original identity signal is lost during retargeting. Whereas prior
  co-speech robot systems primarily evaluate human-motion or rendered-video
  outputs and treat physical embodiment qualitatively
  \cite{yoon2019robots,shen2025ted}, this command-space evaluation directly
  measures how much personalization is preserved after robot embodiment.

  Identity preservation alone is insufficient for interactive deployment; the
  motion must also be generated before the robot begins responding. We define
  \emph{physical response latency} as the time from the end of the user's
  utterance until the NAO begins synchronized speech and gesture. The
  personalization code is computed once during enrollment and reused across
  turns. Once the first 2.27-s LLM/TTS audio window is available, the RTX~4070
  generates its 34-frame personalized gesture in $0.755\pm0.002$\,s, followed by
  $2.23\pm0.08$\,ms for NAO retargeting. Thus,
  \begin{equation}
    L_{\mathrm{onset}} =
    L_{\mathrm{ASR}} + L_{\mathrm{LLM+TTS}}
    + 0.757\,\mathrm{s} + L_{\mathrm{robot}},
  \end{equation}
  where $L_{\mathrm{LLM+TTS}}$ is the time required to produce the first usable
  audio window and $L_{\mathrm{robot}}$ covers network and robot startup. For an
  LLM/TTS latency of 1--2\,s, motion generation is ready after approximately
  1.76--2.76\,s, excluding ASR and robot-command latency. Because each 2.27-s
  window is generated in 0.755\,s, subsequent windows can be prepared during
  playback with a 1.51-s margin. The final four generated frames seed the next
  window to promote smooth temporal transitions.

\subsection{Style-Encoder Selection}
\label{abla-encoder}
\begin{table}[t]
  \centering
  \caption{Encoder identity recognition on unseen BEAT and TED-Ex motion.}
  \label{tab:encoder_selection}
  \small
  \begin{tabular}{l cc}
    \toprule
    Encoder &  BEAT (\%) &  TED-Ex (\%) \\
    \midrule
    TCN (TED-Ex) & $67.4$ & $63.2$ \\
    TCN (BEAT) & $63.6$ & $13.8$ \\
    TCN (combined) & $82.5$ & $62.8$ \\
    ZeroEGGS GRU & $78.5$ & $69.3$ \\
    ZeroEGGS attention & $\mathbf{84.8}$ & $\mathbf{80.5}$ \\
    \bottomrule
  \end{tabular}
\end{table}
At enrollment, the style encoder determines which user information reaches the
generator. It is useful only if clips from the same person produce similar
embeddings while different people remain separable. We test this on real motion
by matching held-out clips to enrollment identities; accuracy measures whether
the embedding retains cues needed for personalization. Table~\ref{tab:encoder_selection}
compares a convolutional sequence encoder (TCN) with recurrent and
attention-based ZeroEGGS encoders on unseen BEAT and TED-Ex identities.
Single-dataset TCNs transfer unevenly; ZeroEGGS attention performs best on both
and is therefore selected.

\subsection{Encoder and Adaptation Ablations}
Our final design first warms up the style encoder with cross-entropy (CE)
identity classification, then jointly refines it with the style adapter under
the diffusion loss while the prior remains frozen. Warm-up separates people;
joint refinement aligns their embeddings with gesture generation.
Table~\ref{tab:encoder_adaptation_ablation} tests whether both stages are needed
by comparing a frozen warm start, scratch deterministic and variational (VAE)
encoders, and our full design. Unlike the preceding encoder selection, this
table evaluates generated motion using SRA and FGD. Per-user LoRA provides an
optimization-based alternative.
\begin{table}[t]
  \centering
  \caption{Five-seed BEAT/Trinity-set ablations. Style guidance $g_s{=}6$.}
  \label{tab:encoder_adaptation_ablation}
  \small
  \begin{tabular}{@{}l cc@{}}
    \toprule
    Variant & SRA (\%) $\uparrow$ & FGD $\downarrow$ \\
    \midrule
    Frozen CE encoder
      & $61.4 \pm 9.3$ & $38.3 \pm 2.9$ \\
    Joint VAE, scratch
      & $29.1 \pm 2.2$ & $43.6 \pm 12.8$ \\
    Joint det., scratch
      & $61.1 \pm 7.4$ & $35.1 \pm 5.2$ \\
    \textbf{Joint det., CE init.\ (ours)}
      & $\mathbf{69.5 \pm 11.1}$ & $\mathbf{34.2 \pm 5.0}$ \\
    \midrule
    Per-user LoRA (60 ep.)
      & $64.4 \pm 6.8$ & $48.4 \pm 46.6$ \\
    \bottomrule
  \end{tabular}
\end{table}

The full design gives the best result: $69.5\%$ SRA and $34.2$ FGD. Freezing
after warm-up gives $61.4\%$ SRA and $38.3$ FGD, so identity classification
alone is insufficient. The scratch deterministic encoder reaches $61.1\%$ SRA,
showing that diffusion training can learn person-specific cues but benefits
from warm-up; the VAE reaches only $29.1\%$. LoRA reaches $64.4\%$ SRA but
requires 60 epochs per user and has unstable FGD ($48.4\pm46.6$; one seed
reaches $131.4$). Thus, adapting lightweight style-conditioning layers while preserving the pretrained diffusion prior is more stable than adapting the generator itself and provides the strongest SRA–FGD trade-off.
\section{Discussion}
Speech and enrollment motion jointly determine gesture identity. The frozen
audio prior exceeds chance SRA on BEAT/Trinity, suggesting that speech carries
identity-related prosody. Yet swapping the embedding under fixed audio redirects
motion toward the embedding owner, while target identity remains recognizable
under synthetic voices. The embedding therefore adds controllable identity
beyond voice.

When these cues disagree, this control has a cost: a wrong-person embedding
raises FGD from $34.2$ to $43.8$ on BEAT/Trinity and from $30.5$ to $49.1$ on
Quest~3. Because audio is fixed, this is consistent with conflict between
voice- and embedding-implied style, although FGD cannot establish the mechanism.
Persona-Gestor derives personalized gesture entirely from raw audio
\cite{zhang2024speech}, whereas Fares et al. and TranSTYLer explicitly
disentangle source content from target style
\cite{fares2023zero,fares2025transtyler}. Together, these findings identify
voice as a style-bearing condition whose interaction with gestural style
remains open.

Embodiment removes further identity information: the real-motion SRA ceiling
falls from $86.7\%$ to $53.8\%$ after NAO retargeting. Quest~3's lower ceiling
and gain may likewise reflect subtler everyday style, although capture hardware
and population are confounded. Finally, our evaluation covers seven
BEAT/Trinity and ten Quest~3 participants without a human perceptual study; SRA
measures recognizable machine-judged identity, not perceptual equivalence.

\section{Conclusion}
We presented an end-to-end system that enrolls an unseen person from about 10
seconds of gesture and personalizes a frozen gestural style prior without per-user
optimization. The embedding improves SRA while preserving FGD on unseen
motion-capture and Quest~3 users, transfers to synthetic speech, and remains
recognizable after NAO retargeting. Together with the Quest~3 capture
application, these results show that a robot can acquire a measurable component
of a new user's gestural identity from a brief enrollment. Human perception and
identity-preserving embodiment remain open challenges.
\bibliographystyle{IEEEtran}
\bibliography{ref}

\end{document}